# 基于话语重写的无监督对话主题分割模型

李奇峰[1]　侯霞[2]

1（北京信息科技大学）北京 100101

2 北京信息科技大学计算机学院 北京 100101

（2022020656@bistu.edu.cn）

**摘　要**　对话主题分割（Dialogue Topic Segmentation, DTS）在各类对话建模任务中扮演着至关重要的角色。以往的 DTS 方法主要关注语义相似性和对话连贯性，以此来评估无监督对话分割的主题相似性。然而，仅凭语义相似性或对话连贯性无法完全识别主题相似性。此外，未标记的对话数据尚未被充分利用。最新的无监督 DTS 方法通过相邻话语匹配和伪分割，从对话数据中学习具有主题感知的话语表示，从而进一步挖掘未标记对话关系中的有用线索。然而，在多轮对话中，话语常常存在共指或省略现象，导致直接使用这些话语进行表示学习可能对相邻话语匹配任务的语义相似度计算产生负面影响。为充分利用对话关系中的有用线索，本研究提出了一种新颖的无监督对话主题分割方法，该方法结合了话语重写（Utterance Rewriting, UR）技术与无监督学习算法，通过重写对话以恢复共指词和省略词，从而有效利用未标记对话中的有用线索。与现有的无监督模型相比，所提出的话语重写主题分割模型（UR-DTS）在主题分割的准确性上显著提升。主要发现在 DialSeg711 上的性能提高约 6%的绝对误差分数和 WD，实现 11.42%的绝对误差分数和 12.97%的 WD。Doc2Dial 是一个更复杂、更大的数据集，其绝对误差分数和 WD 分别提高了大约 3%和 2%，使 SOTA 在绝对误差分数上达到 35.17%，在 WD 上达到 38.49%。这表明该模型在捕捉对话主题的细微差别方面非常有效，同时也表明利用未标记的对话的有用性和挑战性。

**关键词**：话语重写；对话主题分割；无监督；自然语言生成；

**中图法分类号**：

## An Unsupervised Dialogue Topic Segmentation Model Based on Utterance Rewriting

LI Qi-feng[1]，LI [2]

1 (Beijing Information Science & Technology University), Beijing 100101, China

2 School of Computer Science，Beijing Information Science & Technology University，Beijing 100101，China

**Abstract**　Dialogue Topic Segmentation (DTS) plays a crucial role in various types of dialog modeling tasks. Previous DTS methods mainly focus on semantic similarity and dialog coherence to evaluate the topic similarity of unsupervised dialog segmentation. However, semantic similarity or dialog coherence alone cannot fully identify topic similarity. In addition, unlabeled conversation data has not been fully utilized. The state-of-the-art unsupervised DTS methods learn topic-aware discourse representations from conversation data through adjacent discourse matching and pseudo segmentation to further mine useful clues in unlabeled conversational relations. However, in multi-round dialogs, discourses often have co-references or omissions, leading to the fact that direct use of these discourses for representation learning may negatively affect the semantic similarity computation in the neighboring discourse matching task. In order to fully utilize the useful cues in conversational relations, this study proposes a novel unsupervised dialog topic segmentation method that combines the Utterance Rewriting (UR) technique with an unsupervised learning algorithm to efficiently utilize the useful cues in unlabeled dialogs by rewriting the dialogs in order to recover the co-referents and omitted words. Compared with existing unsupervised models, the proposed Discourse Rewriting Topic Segmentation Model (UR-DTS) significantly improves the accuracy of topic segmentation. The main finding is that the performance on DialSeg711 improves by about 6% in absolute error score and WD, achieving an absolute error score of 11.42% and a WD of 12.97%.Doc2Dial, which is a more complex and larger dataset, improves by about 3% in absolute error score and 2% in WD, allowing SOTA to reach 35.17% in absolute error score and WD to 38.49%. This shows that the model is very effective in capturing the nuances of conversation topics, as well as the usefulness and challenges of utilizing unlabeled conversations.

**Keywords**　Utterance Rewriting; Dialogue Topic Segmentation; Unsupervised; Natural Language Generation

# 1 引言

在自然语言处理（Natural Language Process，NLP）领域，对话系统因其在客户服务、个人助理和互动娱乐等不同领域的潜在应用而备受关注。理解和管理对话流是这些系统的重要能力，而对话主题分割（DTS）则是支撑该能力的一项关键技术。

对话主题分割（DTS）作为对话建模的一项基本任务，近年来受到了广泛关注。本质上，DTS 旨在通过将对话分割成主题连贯的部分，以揭示对话的主题结构[1]。对话主题分割对于各种下游对话相关的 NLP 任务起着至关重要的作用，例如对话生成[2]、摘要[3]、响应预测[4]和问答[5]。

近年来，对话主题分割的研究方法集中在有监督学习和无监督学习领域。有监督学习通过标签数据监督对话主题分割，相同数据量的情况下具有更好的分割效果，无监督学习通过聚类和主题建模技术，减少对标注数据的依赖，可用数据集更广泛。

如今对话中的主题分割仍面临着诸多挑战，尤其是在有监督的情况下，标签数据稀缺和不可用。传统的 DTS 方法是采用无监督学习减少对标签数据的依赖，其主要侧重于语义相似性和对话连贯性来评估话题转换。然而，这些方法往往无法全面捕捉话题相似性，因为它们忽略了对话话语中错综复杂的动态变化。此外，由于现有无监督 DTS 方法的局限性，大量未标记的对话数据在很大程度上仍未得到利用。这些方法难以有效利用相邻话语匹配和伪分段，部分原因是多轮对话通常包含共同引用或遗漏，这可能会模糊话语之间的语义关系。这种限制阻碍了无监督模型学习主题感知话语表征的能力，最终影响了主题分割的准确性。

本研究旨在通过开发一种融合了话语重写技术的创新型无监督对话主题分割模型来应对上述挑战。主要目的是增强模型理解和利用对话话语细微差别的能力，从而提高主题分割的准确性。该模型通过重写对话以恢复对话中缺失的信息，充分利用未标记对话中有用的线索。这种方法不仅旨在弥补语义相似性计算方面的差距，还旨在利用各种领域中丰富但未得到充分利用的对话数据。

本文在 Gao 等人[6]工作的基础上进行研究，重点研究如何有效利用未标记的对话数据方面，有两方面的贡献。首先，本文提出了话语重写话题分割模型（UR-DTS），这是一种利用话语重写（UR）技术进行对话话题分割的新型无监督方法。该模型解决了现有无监督 DTS 方法的局限性，尤其是在处理对话中的共同指代和省略方面，是该领域的一大进步。其次，本文对 UR-DTS 模型进行了全面评估，证明与目前最先进的无监督模型相比，UR-DTS 模型在主题分割准确性方面表现出色。研究结果表明，在多个数据集上，UR-DTS 模型在误差分值和 WindowDiff（$W_D$）指标上都有明显改善，凸显了该模型在捕捉复杂对话主题方面的有效性。通过这些贡献，本研究不仅推进了对对话主题分割的理解，还为在对话系统中利用无标记对话数据开辟了新途径。

# 2 相关工作

本章将依次介绍对话主题分割和话语重写技术的现有发展状态，这些方法都已取得良好的效果，对相关领域的发展起到了重要推动作用，对本文的工作也有着很强的借鉴意义。

## 2.1 对话主题分割

对话主题分割（DTS）是将对话会话分割成主题一致的片段的过程，类似于对文档进行分割的任务。在这项任务中，各种最初为文档分割设计的方法都被应用到了基于对话的文本中。最初，由于缺乏训练数据，无监督方法占主导地位。这些方法依赖于分析单词共现统计[7]或检查句子的主题分布[8]来识别对话转折中主题或语义的变化。

随着大规模数据集（如维基百科中的数据集）的出现，对话主题分割的格局开始发生变化，有监督的分割技术开始发展。特别是基于神经网络的技术[9]，因其更高的准确性和效率而广受欢迎。但是，与文档分割相比，对话中语言碎片化、动态化和非正式的用语等情况给对话主题分割造成更多的困难。

同时，由于为训练有监督模型而收集精确注释需要大量费用[10]，而且不同领域的注释指令也不尽相同，因此无监督方法始终是一项研究仍热点。

通常，无监督方法分两个阶段执行。首先，采用各种技术来评估潜在片段分界线（即两个话语之间的间隔）两侧的主题相似性。随后，采用分段算法（如 TextTiling[11]）来识别段落的边界。传统的 DTS 方法主要侧重于通过对话连贯性或通过表面特征（例如词汇重叠）计算的语义相似性[12]来评估主题相似性。近年来，Song 等人[13]和 Xu 等人[4]对 TextTiling 进行了改进，前者结合了词嵌入。后者则使用预训练语言模型[14]对对话进行编码，例如 BERT[15]和 SentenceBERT[16]，以更好地理解对话级的依赖关系，该方法比传统的词袋方法可以更有效的把握语义的细微差别。另外 Xing 等人[17]进一步提出了连贯性评分模型（Coherence Scoring Model，CSM），该模型采用话语对连贯性来评估主题相似性。然而，这些方法往往无法全面捕捉主题相似性，因为它们忽略了对话话语中错综复杂的动态变化。

尽管无监督对话主题分割（DTS）研究取得了重大进展，但仍有一些问题尚未解决，特别是在主题相似性建模和利用无标记对话数据方面。传统方法依赖于一般语义相似性或对话连贯性来确定主题相似性，但这些标准往往存在不足，无法完全捕捉主题相似的同一段落。具体而言，共享同一主题中的语句不一定总是表现出语义相似性，语义相似的话语也可能与同一主题无关。对话连贯性是指话语与其先前上下文之间的响应关系[18]，反映相邻话语是否联系在一起。然而，同一主题片段中的两个不相邻的话语可能在主题上

相似但不连贯。

此外，未标记的对话数据蕴含着丰富的有关对话关系的有用线索，但其潜力仍未得到充分利用。目前的语义相似性方法使用的单词或句子嵌入是在通用文本语料库和监督自然语言推理（NLI）数据集[19]上进行预训练的，这与无标记对话数据的特点不太相符。另一方面，在基于连贯性的方法中，CSM 从 DailyDialog[20]数据集中学习对话连贯性，学习到了一定的对话线索，并且不需要 DTS 任务的注释操作。然而，这些对话中的每一个都只涉及一个主题。因此，CSM 依赖对话级话题标签生成训练样本，并不能完全解决多话题对话中话题分割的复杂性问题。

为了解决上述问题，Gao 等人[6]提出了一种新颖的无监督 DTS 框架，称为具有主题感知话语表示的无监督对话主题分割模型，更进一步的挖掘了未标记对话数据中的有用线索，通过相邻话语匹配（NUM）和伪分割从未标记的对话数据中学习主题感知话语表示，随后将这些主题感知话语表示与对话连贯性结合使用以执行无监督分割。一定程度上缓解了未标记的对话数据的有效利用问题。然而多轮对话通常包含共指和省略[21]的情况，这可能会模糊话语之间的语义关系，同时限制了无监督模型学习主题感知话语表征的能力，最终影响了主题分割的准确性，所以使得更进一步挖掘未标记的对话数据中的有用线索具有挑战性。

## 2.2 话语重写

话语重写任务（Paraphrase Generation or Text Rewriting）是自然语言处理（NLP）领域的一个重要任务，它指的是将一段文本重新表达，使其含义保持不变或基本不变，但表达方式有所改变，也叫句子重写。这种任务在很多应用场景中都非常有用，比如自动摘要[22]、问答系统[23]、机器翻译的后编辑[24]、聊天机器人[25]等。

话语重写任务中恢复的重要对话信息可能对 DTS 的准确性起到积极的作用。最近，其在多个领域引起了广泛关注。在机器翻译中，人们用重写操作来改进 seq2seq 模型的输出生成[26]，Junczys-Dowmunt 等人[27]探索了多种神经架构（CGRU、GRU、M-CGRU 等），这些架构适用于机器翻译输出的自动译后编辑任务。在文本摘要中，重新编辑检索到的候选词可以提供更准确和抽象的摘要，Chen 等人[28]提出了一种准确、快速的总结模型，该模型首先选择突出的句子，然后抽象地重写它们（即压缩和释义），以生成简洁的整体摘要。在对话建模中，Weston 等人[29]将其应用于重写检索模型的输出，类似的工作还有 Rastogi 等人[30]在英语对话中采用了类似的想法，通过重新表述原始话语来简化下游的 SLU（Spoken Language Understanding，SLU）任务。将源输入重写为一些易于处理的标准格式也在信息检索[31]、语义解析[32]或问答[33]中获得了显着的改进，但他们中的大多数没有注意恢复对话中的共指和省略信息[21]，而且只是采用了简单的词典或基于模板的重写策略。对于多轮对话，由于人类语言的复杂性，设计合适的基于模板的重写规则也非常耗时。本文为了很好的适应了多轮对话的复杂性，采用了新的用于一般开放域对话的英语数据集[34]，基于 seq2seq 模型对话语进行重写。

## 3 研究方法

在本节中，我们将介绍提出的方法。从问题表述开始，然后是数据收集和处理。然后，描述模型架构，最后是优化主题感知话语表示和训练目标。

### 3.1 问题表述

对话主题分割旨在识别对话中的片段边界。形式上，给定一个包含 n 个话语的对话 D，即$D=\{u_1,u_2,...,u_n\}$，相邻话语之间有$n-1$个间隔，表示为$V=\{v_1,v_2,...,v_{n-1}\}$。分割算法将片段边界预测为$B=\{b_1,b_2,...,b_k\}$，其中 k 表示边界数，$b_i$表示对话在第 i 个间隔处被划分。

大多数无监督 DTS 方法遵循两阶段范式。首先，对于位于$u_i$和$u_{i+1}$之间的区间$v_i$，计算相关性得分$r_i$。得分越高，区间两侧属于同一片段的可能性就越大。然后，给定相关性得分$R=\{r_1,r_2,...,r_{n-1}\}$，使用分割算法（例如 TextTiling[11]或其派生算法之一）来确定分割边界。以前的方法通常根据一般语义相似性或对话连贯性来评估相关性得分，最新的方法利用从主题感知话语表示中得出的对话连贯性和主题相似性来建模该相关性得分，我们建议优先对话语进行重写，提高主题感知话语表示中主题相似性的语义计算，以优化主题相似性来建模相关性得分。

话语重写属于不完整话语重写的一部分，给定整个对话$D=\{u_1,...,u_{|D|}\}$，我们将上下文定义为$C=\{u_1,...,u_{t-1}\}$，将不完整话语定义为$u_t(t\leq|D|)$。不完整话语重写旨在通过上下文 C 将$u_t$重写为$u_t^*$。重写后的$u_t^*$不仅应具有与$u_t$相同的含义，而且可以单独理解

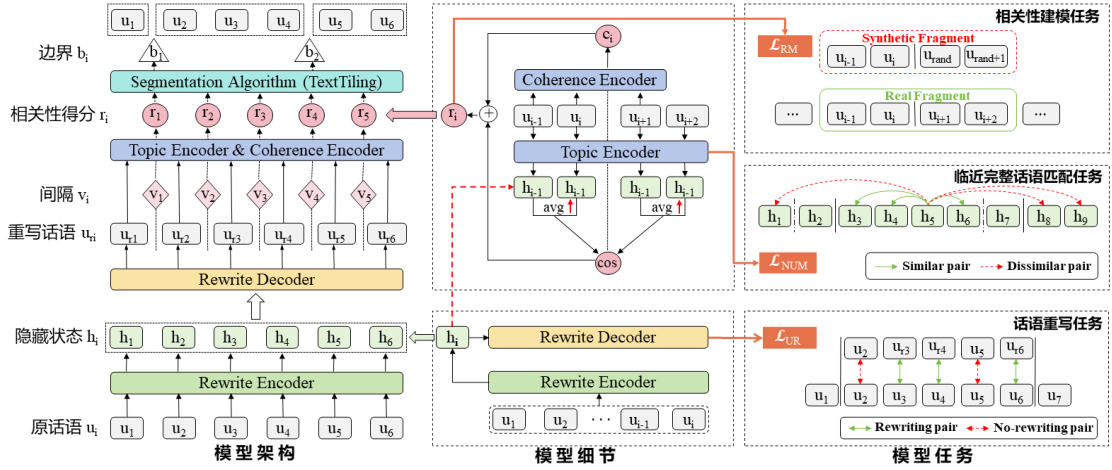

图 1 UR-DTS 框架

### 3.2 模型架构（介绍图内容）（图介绍加上原工作，文字说明，visio）

如图*所示，我们的分割模型由重写编码器、重写解码器、主题编码器、连贯性编码器和分割算法组成。

话语重写模块包含重写编码器和重写解码器，组合为序列到序列(Seq2Seq)任务，其中编码器的第 i 个隐藏状态向量$h_i$，其输入部分的文本包含前文和当前话语的完整段落$(u_1, u_2, …, u_i)$，对应的训练任务是话语重写任务$\mathcal{L}_{UR}$，绿色实线箭头为重写话语对，红色虚线箭头为原话语对。

重写之后的话语为$u_t^*$，相邻话语之间的间隔为$v_i$，为了获得更好的话语表示初始化，我们选择 SimCSE[35]来初始化我们方法的主题编码器。SimCSE 是一个简单但有效的对比句子嵌入框架。我们将重写后的$u_t^*$传入我们的主题编码器以获得每个话语的主题表示：

$$h_i^* = SimCSE(u_i^*)$$

其中$h_i^* \in R^{d_h}$表示 SimCSE 最后一层的池化输出，$d_h$是隐藏状态的维度。按照 CSM[17]，我们选择 Next Sentence Prediction (NSP) BERT[15]作为我们的连贯性编码器。对于每个间隔$v_i$，连贯性编码器计算连贯性得分如下：

$$c_i = NSP - BERT([u_{i-1}^*; u_i^*], u_{i+1}^*)$$

其中$u_{i+1}^*$是当前的响应，$[u_{i-1}^*; u_i^*]$是连接的前文。将多主题多轮对话输入到主题编码器和话语连贯性编码器，在获得主题表示$h_i^*$和连贯性得分$c_i$后，我们计算相关性得分$r_i$为：

$$r_i = sim\left(\frac{h_{i-1}^* + h_i^*}{2}, \frac{h_{i+1}^* + h_{i+2}^*}{2}\right) + c_i$$

其中$sim(.,.)$是余弦相似度，然后相关度得分$r_i$被分割算法用来执行分割。

其中主题编码器对应的训练任务是邻近完整话语匹配任务$\mathcal{L}_{NCUM}$，绿色实线的箭头为定义的主题相似话语对，红色虚线箭头为定义的主题不相似话语对。该任务基于 Gao 等人[6]的工作进行扩展，改用了重写的话语，其恢复了对话中的共指和省略等信息，旨在改进之前工作中计算主题相似性的余弦相似性计算操作，使得模型学习分辨相似的主题的能力得到提升。话语连贯性编码器对应的训练任务是相关性建模任务$\mathcal{L}_{RM}$，绿色框内为真实的连贯片段，红色框内为构建的虚假片段，即不连贯片段。

### 3.3 不完整话语重写

我们将重写视为序列到序列(Seq2Seq)任务，并采用两个预训练的 Seq2Seq 模型，T5[36]和 Pegasus[37]。输入是串联的上下文话语和原始的最后一句话语，每句话语前插入特殊标记以指示其说话者，通过模型的编码器学习原话语的表示，捕捉重要信息：

$$h_i = Enc([CLS] + u_1 + [SEP] + u_2 + … + [SEP] + u_i)$$

其中$u_i$为经过处理的包含上下文的原话语信息，$h_i$表示对应的隐藏状态。

重写解码器接收来自编码器的隐藏状态向量$h_i$，并将其用于生成最终的输出序列。以开始符号和先前生成的词为输入，逐步生成目标序列，同时利用编码器提供的源序列上下文信息，最终输出是一个序列的概率分布序列，指导整个翻译或生成过程。重写解码器的公式如下：

$$Decoder(y) = FFN(Self - Attention(y) + Encoder - Decoder Attention(y, h_i))$$

其中输入向量为$y = (y_1, y_2, …, y_n)$，其中$y_i$是解码器的第 i 个输入。解码器的任务是生成一个输出序列$x = (x_1, x_2, …, x_m)$，其中每个$x_j$是输出的第 j 个元素。

### 3.4 主题感知话语表示

为了训练具有主题感知能力的分割模型，我们基于 Gao 等人[6]的工作进行扩展，提出了一项名为邻近完整话语匹配（NCUM）的新任务。原工作基于主题变化的性质，假设话语更有可能与其邻近话语在主题上相似。为了进一步减少未标记对话中的噪音，结合了邻近话语和伪分割来获得精细的主题相似话语对和不相似话语对。然而原工作只是针对话语间的句子级别的去噪操作，忽略了句子本身内部的噪声，比如话

语中的共指和省略等产生的噪声，于是我们提出了一项名为邻近完整话语匹配（NCUM）的新任务，通过恢复信息后的对话来获得精细的主题相似话语对和不相似话语对，再将两类话语对作为边际排名损失的正样本和负样本。

首先，给定 D 中的重写后的话语 $u_i^*$，我们将其邻近话语索引集 $U_i$ 和非邻近话语索引集 $\bar{U}_i$ 定义为：

$U_i = \{j \in [1,n] \mid w \geq |i-j| \land j \neq i\}$

$\bar{U}_i = \{j \in [1,n] \mid w \geq |i-j|\}$

其中 w 是 $u_i^*$ 前后相邻话语的数量，n 指对话 D 的长度。

在无标签的多主题对话中，来自 NCUM 的监督容易产生相对嘈杂的声音。为了减少噪音，我们基于重写的话语进一步将相邻关系与伪分段相结合，以产生精细的主题相似话语对和不相似话语对。给定重写后的话语 $u_i^*$ 及其伪分段 $segment(i)$，$W_i$ 表示 $segment(i)$ 内的话语索引，$\bar{W}_i$ 表示 $segment(i)$ 外的话语索引：

$W_i = \{j \in [1,n] \mid u_j^* \in segment(u_i^*) \land j \neq i\}$

$\bar{W}_i = \{j \in [1,n] \mid u_j^* \in segment(u_i^*)\}$

基于重写话语 $u_i^*$ 的相邻话语和伪片段，我们获得了 $u_i^*$ 的主题相似话语索引 $P_i^+$ 和主题不相似话语索引 $P_i^-$，如下所示：

$P_i^+ = U_i \cap W_i, \quad P_i^- = \bar{U}_i \cap \bar{W}_i$

其中 $W_i$ 表示公式中得到的主题相似话语索引，$U_i$ 表示公式中得到的邻近话语索引集。

## 4 实验

### 4.1 数据集

本文使用了两类数据集：

（1）用于话语重写的数据：为了获得用于开放域对话话语重写的并行训练数据，我们从 DECODE[34]数据集中抽取了 10000 个对话作为训练集，其是从 DailyDialog[20]和 BST[38]数据集中抽取然后重新标注得来。此外，我们从 DailyDialog 和 BST 中抽取了 800 个对话作为测试集。

（2）用于对话主题分割的数据：采用两个广泛使用的数据集 DialSeg711[4]和 Doc2Dial[39]。DialSeg711是一个真实世界的数据集，包含 711 个英语对话，它结合了两个现有的面向任务的对话数据集 MultiWOZ[40]和 KVRET[41]中的对话。该数据集平均每个对话包含 4.9 个主题段，每个主题段包含 5.6 个话语。Doc2Dial数据集包含 4,100 多个合成英语对话，这些对话基于属于四个领域的 450 多个文档。该数据集平均每个对话包含 3.7 个主题段，每个主题段包含 3.5 个话语。

### 4.2 评估指标

为了确保公平比较，我们采用了两个标准指标，即 $P_k$ 误差分数[42]和 WinDiff(WD)[43]。$P_k$ 和 WD 都是通过测量一定大小的滑动窗口内真实片段与模型预测之间的重叠来计算的，由于它们都是惩罚指标，因此分数越低表示性能越好。

### 4.3 基线

我们将提出的 UR-DTS 模型与以下无监督基线进行比较：

Random：给定一个包含 k 条话语的对话，我们首先随机抽取该对话的片段边界数量 b ∈ {0, ..., k − 1}。然后我们以概率 b k 确定某个话语是否是片段的结尾。

BayesSeg[44]：此方法根据与该段相关的多项式语言模型对每个主题段中的单词进行建模。最大化对话的观察可能性可产生词汇衔接的分段。

GraphSeg[12]：该方法以话语作为节点生成语义相关图。然后通过找到图中的最大团来预测句段。

GreedySeg[4]：该方法根据从预训练的 BERT 句子编码器的输出计算出的相邻话语的相似性，贪婪地确定段边界。

TextTiling[11]：是一种将文本细分为多段落单元的技术，这些单元表示段落或子主题。用于识别主要子主题变化的话语线索是词汇共现和分布的模式。

TeT + Embedding[13]：通过应用词嵌入来计算连续话语对的语义连贯性，通过 GloVe 词嵌入增强TextTiling。

TeT + CLS[4]：通过预训练的 BERT 句子编码器增强 TextTiling，通过使用 BERT 编码器的输出嵌入来计算连续话语对的语义相似度。

TeT + NSP[6]：通过预训练的 BERT 增强了下一句预测(NSP)的 TextTiling，利用输出概率来表示连续话语对的语义一致性。

CSM[17]：通过利用来自话语对连贯性评分任务的监督信号来解决无监督方法仅利用表面特征来评估话语之间的主题连贯性的限制。

　　CSM（unsup）[6]：一种改进的 CSM 变体，其在原本基础上不需要主题标签和动作标签。
　　TeT+ RM + NUM[6]：通过相邻话语匹配和伪分割从未标记的对话数据中学习主题感知话语表示。

### 4.4 实验参数设置

对于话语重写，我们使用了三种预训练模型：T5-Base 和 T5-Large，其参数大小分别为 220M 和 770M。每个模型训练 4 个 epoch，学习率为 5e−5，并使用波束搜索（波束大小为 5）进行生成。

表 1 话语重写实验参数

Table 1 Parameters of the discourse rewriting experiment

| 实验参数 | 值 |
| --- | --- |
| 显卡 | NVIDIA GeForce RTX 3090*1 |
| 数据集-UR | DECODE[34] |
| 学习率 | 5e-5 |
| 批处理大小 | 12 |
| 训练轮次 | 4 |
| 波束 | 5 |

主题编码器我们从 SimCSE 的 sup-simcse-bert-base-uncased 版本的预训练检查点开始。对于 DialSeg711 和 Doc2Dial，我们选择相邻话语的数量 w 为 5，以提高性能和计算效率。

表 2 主题编码器实验参数

Table 2 Theme Encoder Experimental Parameters

| 实验参数 | 值 |
| --- | --- |
| 显卡 | NVIDIA GeForce RTX 3090*1 |
| 数据集-DTS | DialSeg711[4]和 Doc2Dial[39] |
| 学习率 | 1e-5 |
| 批处理大小 | 4 |
| 训练轮次 | 15 |
| 滑动窗口 | 5 |

### 4.5 实验结果

我们将我们提出的方法与两类无监督基线进行了比较：
（1）不使用 TextTiling 的基线，包括 BayesSeg[44]、GraphSeg[12]和 GreedySeg[4]；
（2）从 TextTiling 扩展而来的基线，例如 TeT[11]、TeT+CLS[4]、TeT+Embedding[13]、TeT+NSP[6]、连贯性评分模型（CSM）[17]和 TeT+RM+NUM[6]。其中 CSM 和 TeT+RM+NUM 使用对话连贯性而不是语义相似性，与其他从 TextTiling 扩展而来的无监督基线不同。

表 3 主要实验结果

Table 3 Main experimental results

| Method | DialSeg711 | | Doc2dial | |
| --- | --- | --- | --- | --- |
| | $P_k \downarrow$ | WD $\downarrow$ | $P_k \downarrow$ | WD $\downarrow$ |
| Random | 52.92 | 70.04 | 55.60 | 65.29 |
| BayesSeg[44] | 30.97 | 35.60 | 46.65 | 62.13 |
| GraphSeg[12] | 43.74 | 44.76 | 51.54 | 51.59 |
| GreedySeg[4] | 50.95 | 53.85 | 50.66 | 51.56 |
| TextTiling[11] | 40.44 | 44.63 | 52.02 | 57.42 |
| TeT + Embedding[13] | 39.37 | 41.27 | 53.72 | 55.73 |
| TeT + CLS[4] | 40.49 | 43.14 | 54.34 | 57.92 |
| TeT + NSP[6] | 46.84 | 48.50 | 50.79 | 54.86 |
| CSM[17] | 26.80 | 28.24 | 45.23 | 47.32 |
| CSM（unsup）[6] | 24.30 | 26.35 | 45.30 | 49.84 |
| TeT+ RM + NUM[6] | 17.86 | 19.80 | 38.11 | 40.72 |
| $Ours_{T5-base}$ | 12.00 | 13.35 | 35.97 | 40.01 |
| $Ours_{T5-large}$ | **11.42** | **12.97** | **35.17** | **38.49** |

我们的框架通过恢复话语中关键信息，使用未标记的对话进行训练，另外分段注释仅用于评估。表 3 展示了我们的模型和基线在两个数据集上的结果。我们的模型在两个评估数据集上都实现了最佳(SOTA)性能，与之前的 SOTA 有不同程度的差距。具体来说，我们能够将 DialSeg711 上的性能提高约 6%的绝对误差分数和 WD，实现 11.42%的绝对误差分数和 12.97%的 WD。Doc2Dial 是一个更复杂、更大的数据集，其绝对误差分数和 WD 分别提高了大约 3%和 2%，使 SOTA 在绝对误差分数上达到 35.17%，在 WD 上达到 38.49%。这表明我们的模型通过有效利用了重写之后的未标记对话，恢复了对话中共指和省略等信息，使得模型从学习主题相似性和对话连贯性中获益。

### 4.6 案例研究

表 4 话语重写示例

Table 4 Examples of Discourse Rewriting

| 原话语 | 重写后的话语 |
| --- | --- |
| I need to find a shopping center. | I need to find a shopping center. |
| The Stanford Shopping Center at 773 Alger Dr is 3 miles away in no traffic. Would you like directions there? | The Stanford Shopping Center at 773 Alger Dr is 3 miles away in no traffic. Would you like directions **to the Stanford Shopping Center**? |
| Yes please. | **Yes, I would like directions to the Stanford Shopping Center at 773 Alger Dr, please.** |
| I sent all the info on the screen, please drive carefully! | I sent all the info on the screen, please drive carefully! |
| ================ | ================ |
| Schedule a doctor appointment and my sister is coming along. | Schedule a doctor appointment and my sister is coming along. |
| Okay, when would you like to schedule that for? | Okay, when would you like to schedule **a doctor appointment for your sister**? |
| Schedule it for 4pm today please. | Schedule **my sister's doctor appointment** for 4pm today please. |

**为什么话语重写有帮助？** 如图 4 所示，原话语中的"there、that、it"等共指或省略的信息，通过重写模型得到了还原。基于 Gao 等人[6]的工作进行扩展，其通过邻近话语匹配(NUM)和伪分割从未标记的对话数据中学习主题感知话语表示，有效利用了未标记的对话，但原工作只是针对话语间的句子级别的去噪操作，忽略了句子本身内部的噪声，比如话语中的共指和省略等产生的噪声，为了进一步消除对话中的噪声，恢复同一主题下的关键对话信息，改进模型计算主题相似性的余弦相似性计算操作，由于重写后的相邻话语中共享词汇的增加，提高了主题相似的相邻话语的相似性计算得分，使得主题相似的正例和负例的分数分布更明显，模型学习分辨对话中主题是否相似的能力得到提升。

### 结束语

本文系统地介绍了对话主题分割领域的工作，并提出了基于话语重写的无监督对话主题分割模型，进一步解决了无监督框架下未标记的对话的有效利用问题，并给出了其中算法部分的实验验证，使得我们的模型在两个评估数据集上都实现了最佳(SOTA)性能。

李奇峰，出生于 2000 年，研究生，主要研究方向为自然语言处理。

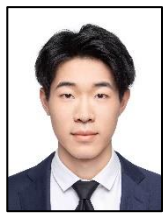

**LI Qi-feng**，born in 2000，postgraduate. Her main research interests include natural language processing (NLP).

# 参考文献